%% file: root.tex
\documentclass[letterpaper, 10 pt, conference]{ieeeconf}  % Comment this line out if you need a4paper

\usepackage{amsmath,graphicx}

\let\labelindent\relax
\usepackage[inline]{enumitem}
\setlist{nosep, leftmargin=14pt}

\usepackage{mwe} % to get dummy images
\usepackage{tabularx} % For tables that span the entire page width
\usepackage{booktabs} % For better table lines
\usepackage{authblk}
\usepackage{multirow}
\title{\LARGE \bf
T2ID-CAS: Diffusion Model and Class Aware Sampling to Mitigate Class Imbalance in Neck Ultrasound Anatomical Landmark Detection }

\author[1]{Manikanta Varaganti}
\author[2]{Amulya Vankayalapati}
\author[2]{Nour Awad}
\author[2]{Gregory R. Dion}
\author[1,3]{Laura J. Brattain}
\affil[1]{Department of Computer Science, University of Central Florida, Orlando, FL, USA}
\affil[2]{Department of Otolaryngology Head Neck Surgery, University of Cincinnati College of Medicine, OH, USA}
\affil[3]{Department of Internal Medicine, University of Central Florida College of Medicine, Orlando, FL, USA}

\begin{document}
%\ninept
%
\maketitle
\begin{abstract}
Neck ultrasound (US) plays a vital role in airway management by providing non-invasive, real-time imaging that enables rapid and precise interventions. Deep learning-based anatomical landmark detection in neck US can further facilitate procedural efficiency. However, class imbalance within datasets, where key structures like tracheal rings and vocal folds are underrepresented, presents significant challenges for object detection models. To address this, we propose T2ID-CAS, a hybrid approach that combines a text-to-image latent diffusion model with class-aware sampling to generate high-quality synthetic samples for underrepresented classes. This approach, rarely explored in the ultrasound domain, improves the representation of minority classes. Experimental results using YOLOv9 for anatomical landmark detection in neck US demonstrated that T2ID-CAS achieved a mean Average Precision of 88.2, significantly surpassing the baseline of 66. This highlights its potential as a computationally efficient and scalable solution for mitigating class imbalance in AI-assisted ultrasound-guided interventions.
\newline

\indent \textit{Clinical relevance}— 
The proposed approach addresses class imbalance by ensuring the accurate detection of underrepresented airway structures like tracheal rings and vocal folds, improving ultrasound-guided airway management. This improves safety, reduces misplacement risks, and supports precise, real-time assessment in critical care, especially for patients with difficult airway anatomy.
\end{abstract}

\input{Text/1_introduction}

\input{Text/2_methods}

\input{Text/3_experiments}

\input{Text/4_results_discussions}
\input{Text/5_conclusion}
\input{Text/6_compliance_ethical_standard}
\bibliographystyle{IEEEbib}
\bibliography{Bib/refs, Bib/strings}

\end{document}

%% file: Text/1_introduction.tex
\section{Introduction}
\label{sec:intro}
The assessment of anatomical landmarks in neck ultrasound (US) is crucial for securing the airway, particularly in emergency situations that require rapid airway management \cite{osman2016role}. US imaging helps physicians precisely place endotracheal tubes and ensures that they are positioned correctly by visualizing important structures, such as the thyroid cartilage, strap muscles, tracheal rings, and thyroid gland \cite{vsustic2007role}. However, the identification and tracking of these anatomical features is hampered by the high variability of patient body types and operator skill levels.   

 Deep learning based object detection models have shown remarkable success in automating the identification of anatomical structures in medical imaging. However, their performance is often hindered by class imbalance \cite{9042296}, a common issue in medical datasets in which certain structures are underrepresented. This imbalance leads to biased models that prioritize majority classes, resulting in poor detection rates for minority classes, which are often clinically significant. Addressing this imbalance is necessary to enhance detection accuracy and ensure that object detection algorithms can be effectively generalized across all classes.  
 
Traditional approaches to addressing class imbalance, such as oversampling the minority class \cite{chawla2002smote} or undersampling the majority class \cite{drummond2003c4}, come with inherent limitations \cite{8125820}. Oversampling can increase the risk of overfitting, while undersampling may lead to the loss of critical information from the majority class. Standard data augmentation techniques, such as mosaic augmentation \cite{bochkovskiy2020yolov4}, introduced in YOLOv4, combine multiple images to enhance contextual diversity, while mixup \cite{zhang2017mixup} interpolates between image-label pairs to improve model generalization. However, these methods have not been extensively evaluated in medical imaging, where maintaining anatomical accuracy presents unique challenges. Class-aware sampling (CAS) addresses class imbalance by adjusting the sampling probability of underrepresented classes to create balanced mini-batches during training \cite{shen2016relay}. Similarly, Repeat Factor Sampling (RFS) mitigates class imbalance by duplicating images containing rare classes based on their inverse frequency \cite{gupta2019lvis}.

\begin{figure*}[h]
  \centering
  \centerline{\includegraphics[width=18cm]{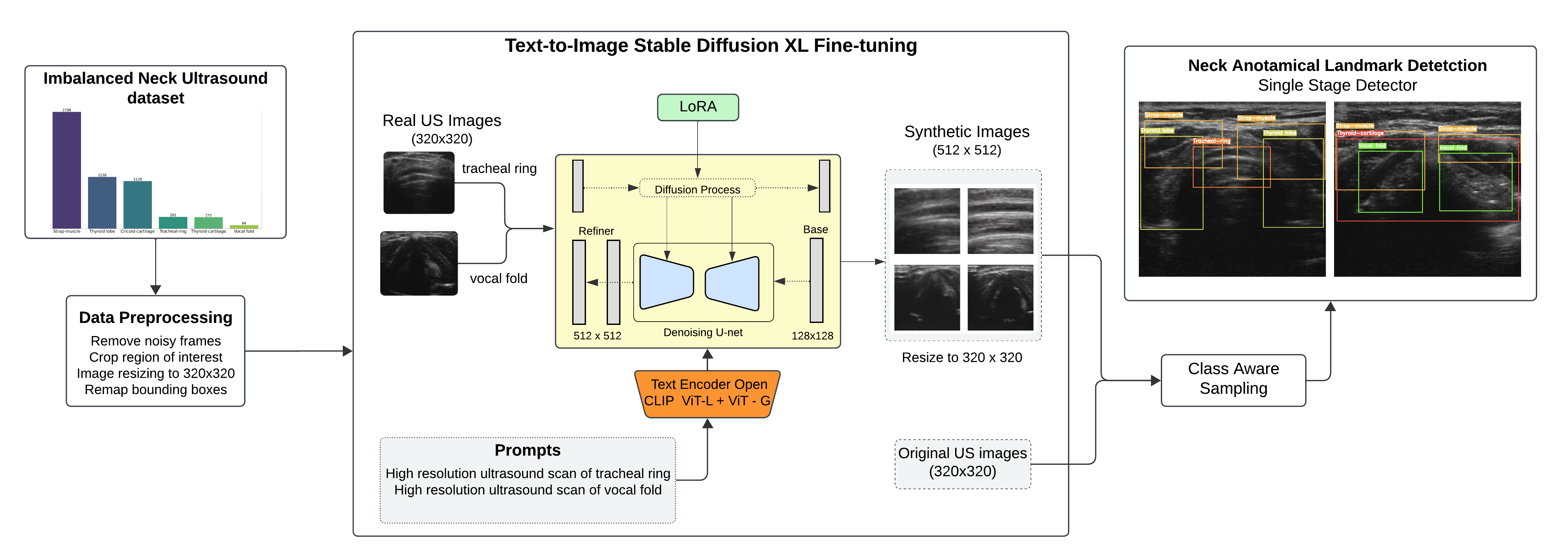}}
%  \vspace{1.5cm}
\caption{\small Overview of the proposed T2ID-CAS framework}
\label{fig:T2IDCAS}
\end{figure*}

Synthetic data generation helps address class imbalance by creating artificial samples for minority classes \cite{ye2023exploiting}. Techniques like the Synthetic Minority Over-sampling Technique (SMOTE) \cite{chawla2002smote} and Generative Adversarial Networks (GANs) \cite{goodfellow2014generative} are commonly used in natural image datasets but face challenges in medical imaging due to the requirement for large datasets. Diffusion models present a promising alternative, capable of generating high-fidelity synthetic images by progressively adding Gaussian noise to data and then reversing the process through learned denoising steps, transforming random noise into coherent images \cite{ho2020denoisingdiffusionprobabilisticmodels}. These models have demonstrated success across various medical imaging modalities. For instance, Lei et al. \cite{lei2024modalities} generated high-quality 3D brain MRI images, while Chung et al. \cite{chung2022score} employed score-based diffusion methods to enhance artifact suppression and contrast in MRI reconstructions. Additionally, Xu et al. \cite{xu2024medsyn} utilized diffusion models to produce high-quality 3D lung CT images guided by textual information. Despite the growing adoption of diffusion models in medical imaging, their application in ultrasound remains limited.

In this study, we introduce T2ID-CAS, a novel Text-to-Image Diffusion model with Class-Aware Sampling (CAS) designed to address class imbalance in anatomical landmark detection for neck ultrasound. Our key contributions include: (i) fine-tuning a text-to-image stable diffusion model on ultrasound images of tracheal rings and vocal folds to accurately capture their anatomical features, (ii) constructing a hybrid dataset by combining real and synthetic images to achieve balanced class representation, and (iii) integrating Low-Rank Adaptation (LoRA) to enhance the computational efficiency of stable diffusion. Through extensive experiments, we demonstrate that our approach significantly improves detection performance across all classes, particularly enhancing the mean average precision (mAP50-95) for underrepresented anatomical landmarks. These findings underscore the potential of diffusion-based augmentation in medical ultrasound imaging and position T2ID-CAS as a scalable, computationally efficient solution for mitigating class imbalance in deep learning models.

%% file: Text/2_methods.tex
\section{Proposed Methods}
\label{sec:method}
Our proposed framework T2ID-CAS addresses class imbalance in neck ultrasound imaging by integrating text-to-image latent diffusion based synthetic data generation with class-aware sampling (CAS), augmenting the representation of underrepresented anatomical structures. The framework comprises following core components: (i) a text-to-image latent diffusion model for synthetic data generation, (ii) CAS to prioritize underrepresented classes, (iii) Low-Rank Adaptation (LoRA), and (iv) YOLOv9 based object detection model for neck anatomical landmark detection. The workflow is illustrated in Fig.~\ref{fig:T2IDCAS} and described in detail below.

\subsection{Text-to-Image Latent Diffusion Model}
\label{ssec:data_aug_diffusion}
Diffusion models are generative models that add noise to data (forward diffusion) and learn to reverse this process to generate new samples (reverse diffusion) \cite{ho2020denoising}. Stable Diffusion is a latent diffusion model that operates in a compressed latent space, thereby reducing computational costs \cite{rombach2022high}. These models are designed to mitigate challenges such as training instability and mode collapse, which are commonly observed in standard diffusion models. The architecture comprises of  a variational autoencoder (VAE) which compresses images into a lower-dimensional latent space \cite{kingma2013auto},  denoising U-Net for reversing the diffusion on the latent vectors \cite{ho2020denoisingdiffusionprobabilisticmodels} , and CLIP ViT-L/14 text encoder  for prompt conditioning \cite{radford2021learning}. 

SDXL is a pretrained Stable Diffusion XL model that  extends the original stable diffusion architecture with a larger U-Net backbone and cross-attention layers, thereby enabling higher-resolution image synthesis \cite{podell2023sdxl}. Fine-tuning large-scale diffusion models such as SDXL for medical imaging is computationally expensive and impractical due to limited annotated datasets and high hardware requirements. LoRA \cite{hu2021lora} is a parameter-efficient finetuning method that injects lightweight trainable layers into the model's attention blocks, allowing domain-specific learning without modifying the entire model. This approach has been shown to improve the stability and performance in text-to-image generation tasks in medical imaging applications \cite{farooq2024derm} where data availability is limited and computational efficiency is critical. In our study, we fine-tuned SDXL on real ultrasound images of underrepresented classes (i.e., tracheal rings and vocal folds) and used LoRA for optimal fine-tuning without fully retraining the model.   

After fine-tuning, the model is conditioned on class-specific textual prompts to generate anatomically accurate synthetic images. We used both qualitative and quantitative metrics to assess the quality and diversity of these images. Qualitative evaluation involved human assessment of image-text alignment, visual quality, and fidelity. For quantitative analysis, we measured synthetic image quality using CLIP Score, Inception Score (IS), and Fréchet Inception Distance (FID). CLIP Score evaluates image-text alignment to ensure semantic accuracy \cite{hessel2021clipscore}, while IS assesses image quality and diversity, with higher scores indicating greater variability \cite{salimans2016improved}. FID measures realism by comparing the feature distributions of real and synthetic images, where lower values indicate higher fidelity \cite{heusel2017gans}. These metrics collectively ensure that the generated images are realistic, diverse, and accurately aligned with anatomical structures.

\subsection{Class-Aware Sampling}
\label{ssec:cfs}
The class-aware sampling (CAS) is an effective technique designed to mitigate class imbalance in object-detection datasets. Originally proposed in previous works \cite{shen2016relay}, \cite{shu2023cmw}, CAS optimizes the composition of minibatches during the training process, ensuring that each class is represented uniformly, regardless of its prevalence in the dataset. Our CAS approach maintains a list of anatomical classes relevant to our neck US images, each paired with a corresponding set of images. During training, the algorithm randomly selects a class and then chooses an image containing an instance of that class to construct mini batches. This method guarantees that each class, regardless of its frequency of occurrence in the data set, has the same opportunity to be represented in training batches. CAS enhances the model's exposure to minority classes, thereby improving its ability to generalize across diverse anatomical structures.

\subsection{Neck anatomical landmark detection}
\label{ssec:model_architecture}
Object detection models can be categorized into two main types: two-stage and single-stage detectors. Two-stage detectors, such as Faster R-CNN \cite{ren2015faster}, operate in a two-step process where the first stage generates region proposals and the second stage classifies and refines these proposals. In contrast, single-stage detectors \cite{liu2016ssd}, \cite{redmon2016you} directly predict bounding boxes and class probabilities from the input image in a single pass, providing faster real-time performance.   

The evolution of YOLO models has improved both the speed and accuracy. From the original YOLO to versions such as YOLOv2, YOLOv3, and YOLOv5, features such as multiscale predictions and improved handling of small objects have been integrated to satisfy the demands of real-time object detection. YOLOv9 \cite{wang2024yolov9}, one of the latest in this series, combines Programmable Gradient Information (PGI) and a Generalized Efficient Layer Aggregation Network (GELAN) to achieve remarkable improvements in efficiency, accuracy, and adaptability.  We adopted the YOLOv9 small (YOLOv9s) model\footnote{https://github.com/ultralytics/ultralytics}, which offers the best trade-off between the parameter efficiency and computational load in our experiments. We combine the original images with CAS and synthetic images generated by the SDXL model to finetune the YOLOv9s model. As a comparison, we also trained a model from scratch on neck US images, serving as the baseline for evaluating our approaches to address the class imbalance.

%% file: Text/3_experiments.tex
\section{Materials and Experiments}
\label{sec:implementation}
\subsection{Data Description}
\label{ssec:dataset}

\begin{figure}[htb]
\begin{minipage}[b]{1.0\linewidth}
  \centering
  \centerline{\includegraphics[width=8.6cm]{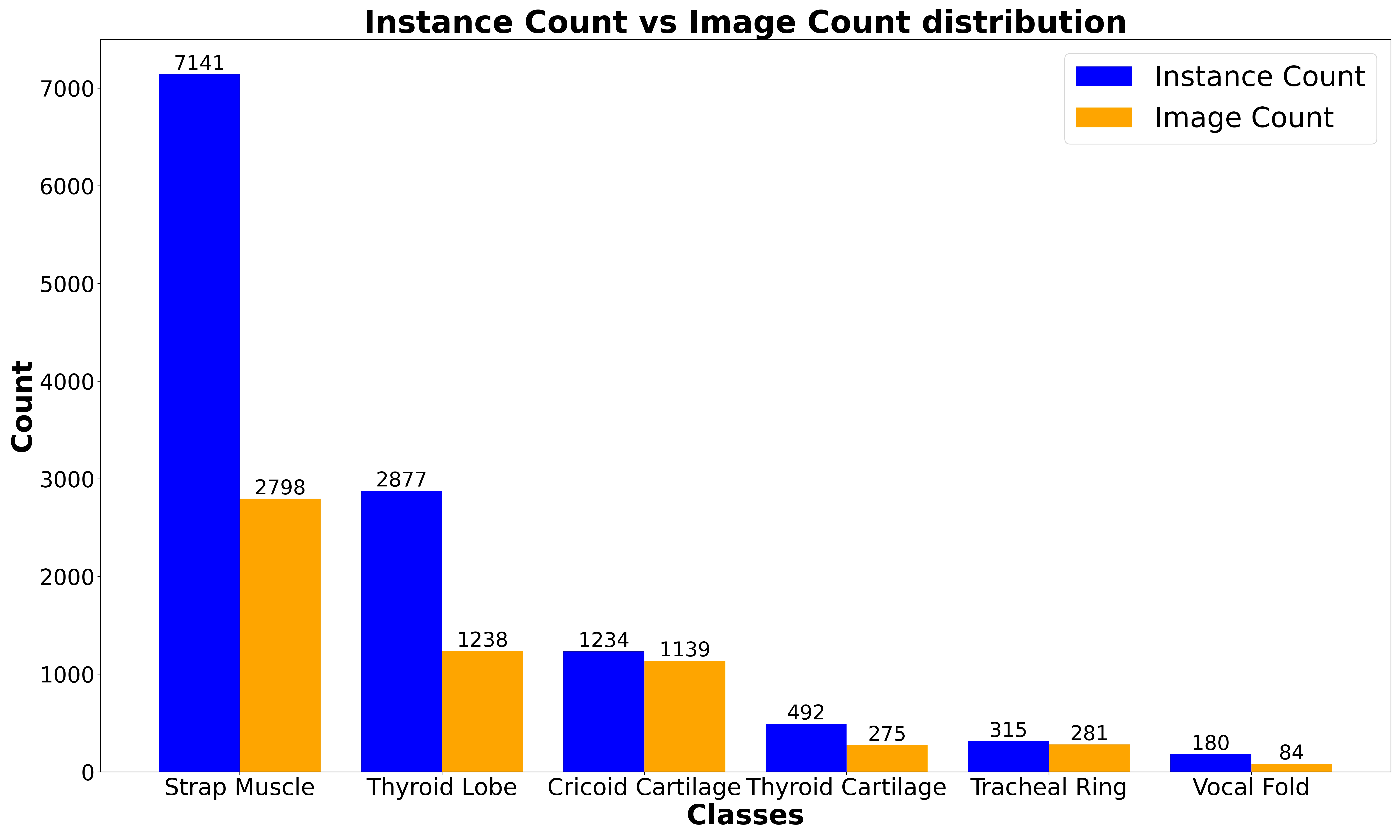}}
%  \vspace{2.0cm}
\end{minipage}
\caption{Long-tailed distribution of instance and image counts per class in the neck US dataset. Colored bars indicate the number of annotated instances (blue) and the number of images (orange) per class.}
\label{fig:longtail}
\end{figure}

Neck US images were collected from 10 human subjects (3 Male/7 Female, Average Age  52.6 ± 14.5) using a Terason uSmart 3200t US device (Teratech, Burlington, MA) with a 15L4A linear probe at a 4 cm imaging depth. To simulate airway assessment, four 10-second cineloops were collected from each subject with a US probe in the short axis, transverse view, moving cranial-caudally from the hyoid bone to the substernal notch, medial-laterally at the tracheal rings, and counter-clockwise rotation around the tracheal rings. The US scans were annotated by airway specialists using YoloMark \cite{bochkovskiy2020yolov4} to generate bounding boxes and class labels for model training. The dataset included 7464 images of six classes: thyroid cartilage, strap muscles, cricoid cartilage, thyroid lobes, vocal folds, and tracheal rings.  Owing to the differences in length, anatomical location, and appearance in US, class distribution is highly skewed. The dataset suffers from a long-tailed distribution, with certain classes (e.g., tracheal rings and vocal folds) being significantly underrepresented (Fig. ~\ref{fig:longtail}). Raw ultrasound images were first cleaned by removing frames with excessive noise and artifacts. Subsequently, the regions of interest containing the relevant anatomical structures were cropped, eliminating extraneous background data. 
The selected images were then uniformly resized to a resolution of 320×320 pixels, and the bounding box coordinates were remapped to reflect the new regions of interest.

\subsection{Training Configuration}
\label{training_setup}
We fine-tuned the SDXL 1.0 model with LoRA\footnote{https://github.com/huggingface/diffusers} with a training data of 840 real US images paired with anatomical text prompts for tracheal rings and vocal folds. The prompts include: "High-resolution ultrasound scan of the human tracheal ring" and "High-resolution ultrasound scan of the human vocal fold". The model was trained for 100 epochs with a batch size of 16, a learning rate of 0.0001, and mixed-precision (fp16) acceleration. Random horizontal flips were applied during fine-tuning for data augmentation, and checkpoints were saved every 5000 steps.

The YOLOv9s model training configuration included the AdamW optimizer with a weight decay of 0.01 and an initial learning rate of 0.001. The input resolution was set to 320×320 pixels, with a batch size of 64 and a dropout rate of 0.1 for regularization. The dataset was split into training and validation sets, with 4-fold cross-validation to ensure a robust model. All the training, inference and evaluations were conducted on an NVIDIA H100 PCIe GPU with 81 GB dedicated memory.

\subsection{Image Generation with Stable Diffusion XL}
\label{ssec:stable_diffusion_image_generation}
\begin{figure}[htb]
\begin{minipage}[b]{1.0\linewidth}
  \centering
  \centerline{\includegraphics[width=9cm]{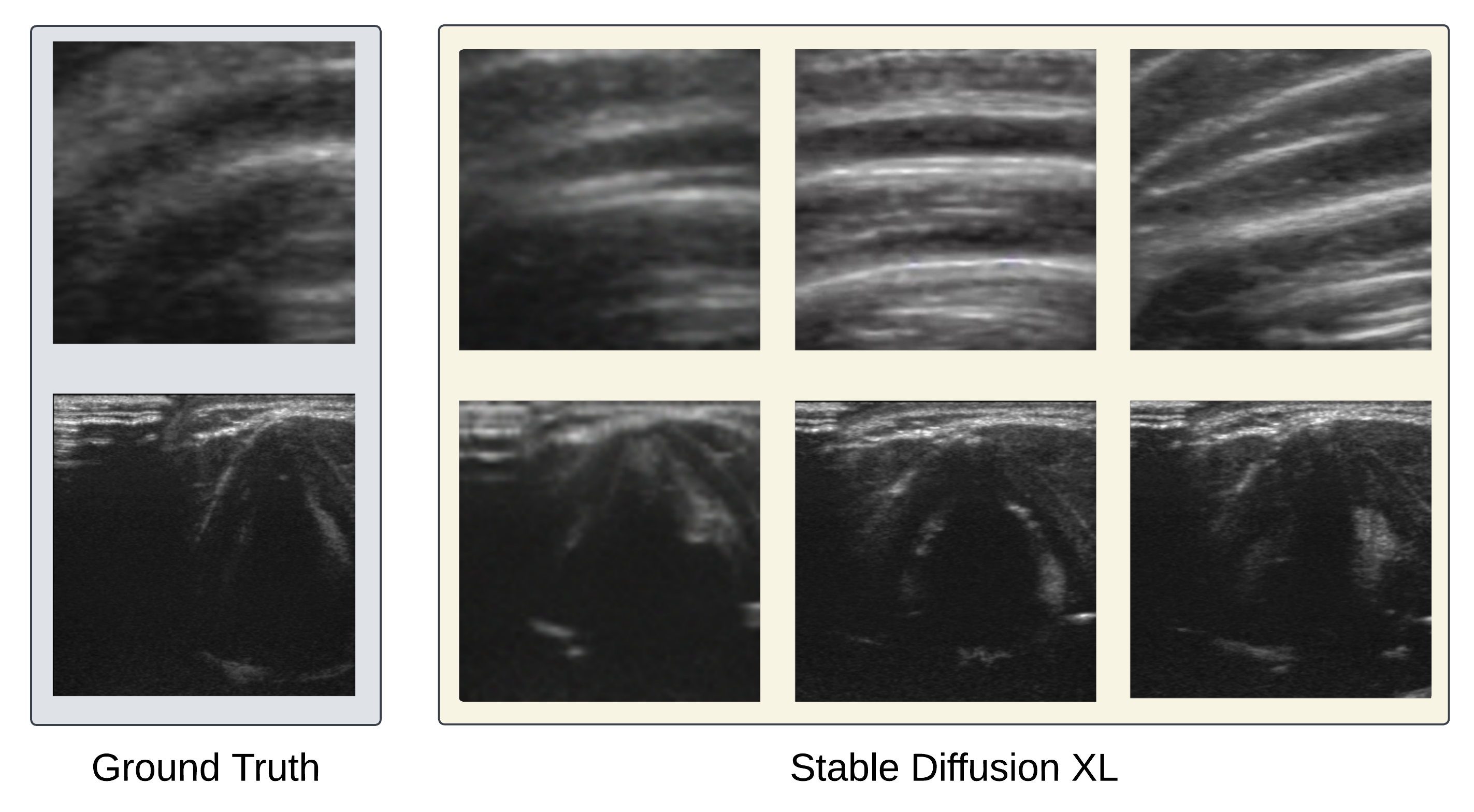}}
%  \vspace{1.5cm}
\end{minipage}
\caption{\small Comparison between original images and synthetic images by SDXL.
(Top) shows the results of tracheal ring using prompt "Ultrasound image of human tracheal ring". (Bottom) shows the results of vocal fold using prompt "Ultrasound image of human vocal fold"}
\label{fig:diffusion}
\end{figure}

Once SDXL was fine-tuned on ultrasound images, it was used to generate high-fidelity synthetic images for tracheal rings and vocal folds. Synthetic images were conditioned on anatomical text prompts with the classifier free guidance (allowed for more precise control over the image generation process) configured to a value of 7.5. Fig. ~\ref{fig:diffusion} shows the results along with user text prompts.  A total of 600 synthetic images of size 512 x 512 were generated, with 300 images for each of the classes of tracheal rings and vocal folds. We evaluated image quality using human assessment and CLIP score, IS and FID. These synthetic images were initially generated at 512×512 pixels to capture finer anatomical details. However, for model training, both real and synthetic images were uniformly resized to 320×320 pixels to ensure computational efficiency and maintain consistency for the YOLOv9 model.

\subsection{Comparative Experiments}
\label{ssec:experiments}
To evaluate the efficacy of our approach, we conducted a series of comparative experiments, assessing different strategies for mitigating class imbalance in neck US anatomical landmark detection. The following experimental configurations were evaluated: baseline (without any sampling), standard data augmentation (mosaic and mixup), Repeat Factor Sampling (RFS), CAS, baseline with SDXL Data, RFS with SDXL Data, and T2ID-CAS(Ours). Performance of YOLOv9s was measured using mean Average Precision (mAP50-95), capturing detection accuracy across multiple IoU thresholds. Additionally, per-class mAP50-95 was analyzed to assess improvements in detecting underrepresented classes.

%% file: Text/4_results_discussions.tex
\section{Results and Discussions}
\label{sec:results}
\subsection{Stable Diffusion XL results}
The synthetic images generated by SDXL model for the tracheal ring and vocal fold were evaluated using FID, IS and CLIP score. Additionally, we compared SDXL against Stable Diffusion\footnote{https://github.com/CompVis/stable-diffusion}(SD)v1-4 using the same textual prompts to ensure a fair comparison. The results tabulated in Table ~\ref{tab:sdxl_results} demonstrate that synthetic images effectively enhanced the representation of underrepresented classes in the dataset. Experimental results indicated that for the vocal fold class, SDXL-LoRA achieves a lower FID (5.54 vs. 9.55), indicating improved realism, and a higher IS (17.623 vs. 17.129), suggesting a greater image diversity. The CLIP Score also improved from 26.3112 to 29.8079, reflecting better semantic consistency with the textual prompts. A similar trend is observed for the tracheal ring class, where SDXL-LoRA achieves an FID of 16.112 compared to 18.12 for SD v1-4, along with notable improvements in IS (18.184 vs. 17.045) and CLIP Score (30.4036 vs. 28.664). These findings indicate that SDXL model, particularly when fine-tuned with LoRA, can effectively generate synthetic ultrasound images that augment class representation for underrepresented anatomical structures.

\begin{table}[h]
\vspace{3pt}
    \centering
    \caption{Quantitative results for Stable Diffusion XL using LoRA}
    \begin{minipage}{\columnwidth}
        \input{Tables/SDXL_metrics}
    \end{minipage}
\end{table}

\begin{figure}[htb]
\begin{minipage}[b]{1.0\linewidth}
  \centering
  \centerline{\includegraphics[width=8.5cm]{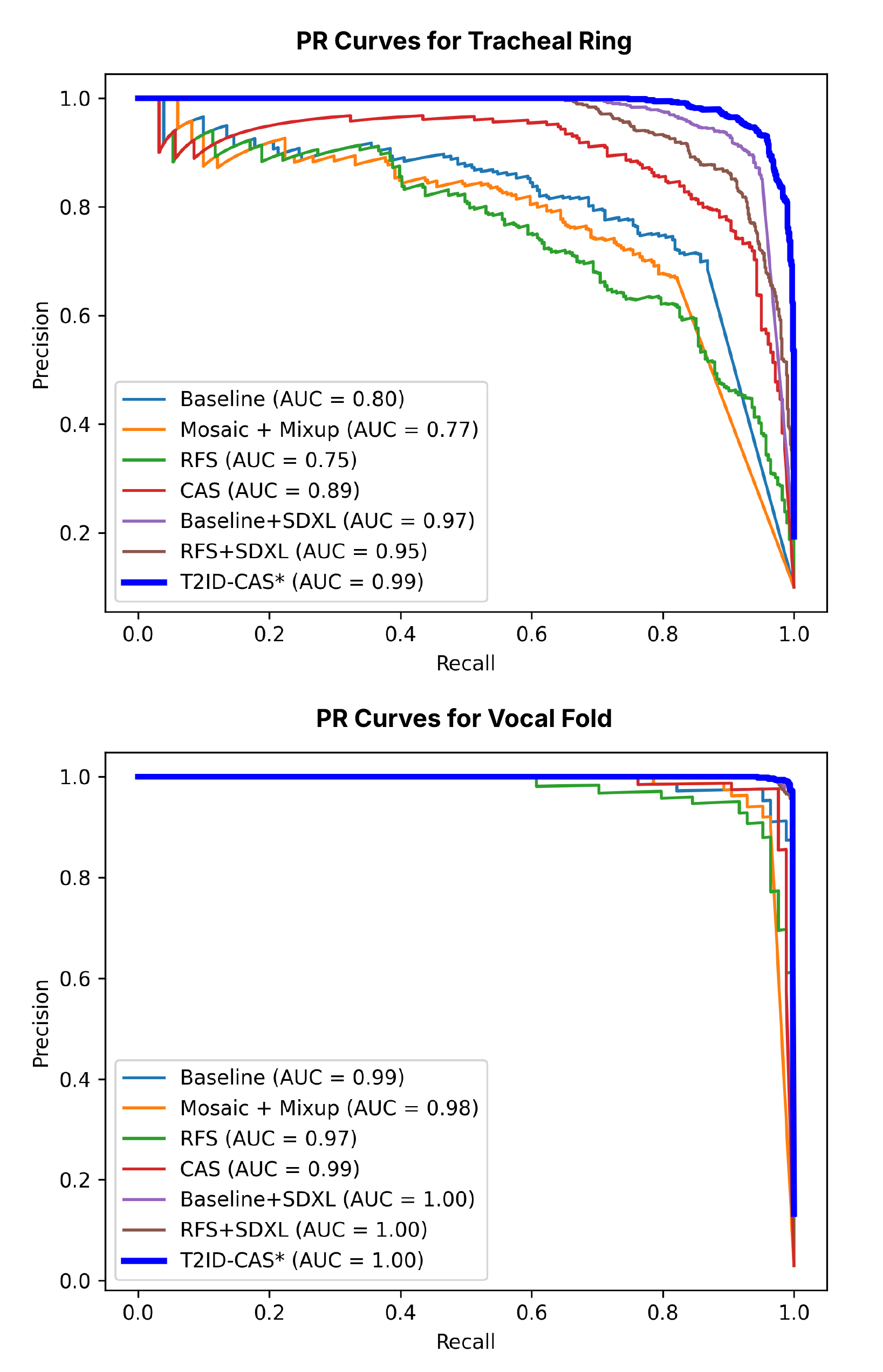}}
%  \vspace{1.5cm}
\end{minipage}
\caption{\small{Precision-Recall curves for Tracheal Ring (Top) and Vocal Fold (Bottom) across different configurations.}}
\label{fig:pr}
\end{figure}

\input{Tables/mAP_metrics}

\subsection{YOLOv9s results}
The YOLOv9s model performance was assessed using mAP50-95 (expressed as a percentage \%) for all classes in different experiments. The baseline model was used as the reference.  The experimental results are summarized in Table ~\ref{tab:experiment_res}. The results revealed consistent improvements in the CAS, achieving an overall mAP50-95 of 84.3, significantly outperformed the baseline value of 66. Our approach T2ID-CAS was further improved to an mAP50-95 of 88.2, indicating that the integration of synthetic images improved the generalizability of the model. SDXL synthetic image images were particularly effective in underrepresented classes, such as the tracheal ring and vocal folds (Fig. ~\ref{fig:pr}). For the tracheal ring, the baseline mAP50-95 was 38.5, but this increased substantially to 82.1 with the introduction of synthetic data and reached 90.5 when using our approach. Similarly, the vocal fold class showed a significant boost achieving an mAP50-95 of 98.2. However, RFS and its variants were less effective, with a maximum mAP50-95 of 74.9 when combined with SDXL results.

\subsection{Discussions}
\label{ssec:discussions}
These findings demonstrate that T2ID-CAS effectively mitigates class imbalance by promoting a more uniform representation of all classes during training. Our approach also improves the model’s generalization ability, particularly for minority classes. This suggests that combining strategic data augmentation with appropriate sampling mechanisms can significantly improve deep learning detection performance for sparsely represented classes.

While the integration of synthetic data generally increases model accuracy, it occasionally leads to minor performance declines for well-represented classes such as the thyroid and cricoid cartilage. This indicates that while synthetic data generation benefits underrepresented classes, it may introduce noise to already well-represented ones, potentially causing slight regression.

Our method does have some limitations. Fine-tuning SDXL remains computationally expensive, even with the integration of LoRA. Additionally, the quality of generated synthetic images depends heavily on the precision of textual prompts. In the future, we plan to test our approach on larger datasets to further enhance the robustness of our results. We will investigate techniques to improve the computational efficiency of fine-tuning and study the model sensitivity to textural prompts.

%% file: Tables/SDXL_metrics.tex
\begin{tabular}{|c|c|c|c|c|}
    \hline
    \textbf{Model} & \textbf{Class} & \textbf{FID $\downarrow$ } & \textbf{IS $\uparrow$ } & \textbf{CLIP Score $\uparrow$} \\
    \hline
    \multirow{2}{*}{SD v1-4} & Vocal fold & 9.55 & 17.129 & 26.3112 \\
    \cline{2-5}
    & Tracheal ring & 18.12 & 11.045 & 28.664 \\
    \hline
        \multirow{2}{*}{SDXL-LoRA} & Vocal fold & 5.54 & 17.623 & 29.8079 \\
    \cline{2-5}
    & Tracheal ring & 16.112 & 18.184 & 30.4036 \\
    \hline
\end{tabular}
\label{tab:sdxl_results}

%% file: Tables/mAP_metrics.tex
\begin{table*}[t]
\caption{Performance metrics of different strategies evaluated on YOLOv9s}
    \centering
    \begin{tabular}{|c| c| c| c| c| c| c| c|}
        \hline
        \textbf{Metric (mAP50-95)} & \textbf{Baseline\% } &  \textbf{Mosaic+Mixup\%} & \textbf{RFS\%} & \textbf{CAS\%} & \textbf{Baseline+SDXL\%} & \textbf{RFS+SDXL\%} & \textbf{T2ID-CAS(Ours)\%} \\
        \hline

         All classes & 66 &66.5 & 65.7 & \textbf{84.3} & 75.2 & 74.9 & \textbf{88.2} \\
        \hline
        Thyroid cartilage & 79.6 &80.5 & 83.4 & \textbf{96} & 74 & 75.1 & 93.5 \\
        \hline
        Cricoid cartilage & 72.4 & 73.4 & 71.1 & 86.2 & 70.8 & 70.2 & 83.9 \\
        \hline
        Tracheal ring & 38.5 & 36.9 & 37.1 & \textbf{63.4} & \textbf{82.1} & 81 & \textbf{90.5} \\
        \hline
        Strap muscle & 65.3 & 67.1 & 65 & 81.9 & 65 & 64.7 & 79.6 \\
        \hline
        Thyroid lobe & 64.6 & 66.9 & 62.4 & \textbf{83.5} & 64.9 & 63.7 & \textbf{83.6} \\
        \hline
        Vocal fold & 75.6 & 74.2 &74.9 & \textbf{95} & 94.6 & 94.7 & \textbf{98.2} \\
        \hline
    \end{tabular}
    
    \label{tab:experiment_res}
\end{table*}

%% file: Text/5_conclusion.tex
\section{Conclusion}
\label{sec:conclusion}
In this study, we focused on mitigating class imbalance in deep learning-based neck US anatomical landmark detection, which is crucial for effective airway management. Our proposed approach, T2ID-CAS, strategically enhances the representation of underrepresented structures, such as tracheal rings and vocal folds, by generating high-quality synthetic images using a class aware sampling technique combined with text-to-image stable diffusion model. We also utilized LoRA to significantly reduce the computing resources needed. We evaluated several strategies including standard data augmentation (mosaic and mixup), CAS, and RFS. Our findings show that T2ID-CAS significantly improves the mAP50-95 across all classes compared to the baseline. While CAS effectively balances class distributions, its integration with Stable Diffusion XL (SDXL) synthetic images further improves model generalization, particularly for minority classes. The  demonstrated improvement in detection accuracy for underrepresented anatomical landmarks underscores its potential to enhance the reliability of ultrasound-guided procedures for rapid and accurate airway management. 

%% file: Text/6_compliance_ethical_standard.tex
\section{Compliance with ethical standards}
\label{sec:ethics}
Study was conducted in accordance with the principles of the Declaration of Helsinki. All data were collected and processed according to ethical protocols approved by the Institutional Review Board.